\documentclass[default]{sn-jnl}% Default

\jyear{2022}%

\theoremstyle{thmstyleone}%
\theoremstyle{thmstyletwo}%

\theoremstyle{thmstylethree}%
\begin{document}

\title[Article Title]{A Lightweight Reconstruction Network for Surface Defect Inspection}

%%=============================================================%%
%% Prefix   -> \pfx{Dr}
%% GivenName    -> \fnm{Joergen W.}
%% Particle -> \spfx{van der} -> surname prefix
%% FamilyName   -> \sur{Ploeg}
%% Suffix   -> \sfx{IV}
%% NatureName   -> \tanm{Poet Laureate} -> Title after name
%% Degrees  -> \dgr{MSc, PhD}

\author{\fnm{Chao} \sur{Hu}}
\author{\fnm{Jian} \sur{Yao}}
\author{\fnm{Weijie} \sur{Wu}}
\author{\fnm{Weibin} \sur{Qiu}}
\author{\fnm{Liqiang} \sur{Zhu}}

\affil{\orgdiv{AI Lab}, \orgname{Unicom (Shanghai) Industry Internet Co., Ltd.}, \orgaddress{\city{Shanghai} \postcode{200000},  \country{China}}}

\abstract{Currently, most deep learning methods cannot solve the problem of scarcity of industrial product defect samples and significant differences in characteristics. This paper proposes an unsupervised defect detection algorithm based on a reconstruction network, which is realized using only a large number of easily obtained defect-free sample data. The network includes two parts: image reconstruction and surface defect area detection. The reconstruction network is designed through a fully convolutional autoencoder with a lightweight structure. Only a small number of normal samples are used for training so that the reconstruction network can be A defect-free reconstructed image is generated. A function combining structural loss and $\mathit{L}1$ loss is proposed as the loss function of the reconstruction network to solve the problem of poor detection of irregular texture surface defects. Further, the residual of the reconstructed image and the image to be tested is used as the possible region of the defect, and conventional image operations can realize the location of the fault. The unsupervised defect detection algorithm of the proposed reconstruction network is used on multiple defect image sample sets. Compared with other similar algorithms, the results show that the unsupervised defect detection algorithm of the reconstructed network has strong robustness and accuracy.}

\keywords{Auto encoder, surface defects, abnormal defects, visual inspection, unsupervised defect}

\maketitle

\section{Introduction}\label{sec1}
\begin{figure}[htbp]
	\centering
     \includegraphics[width=0.95\textwidth]{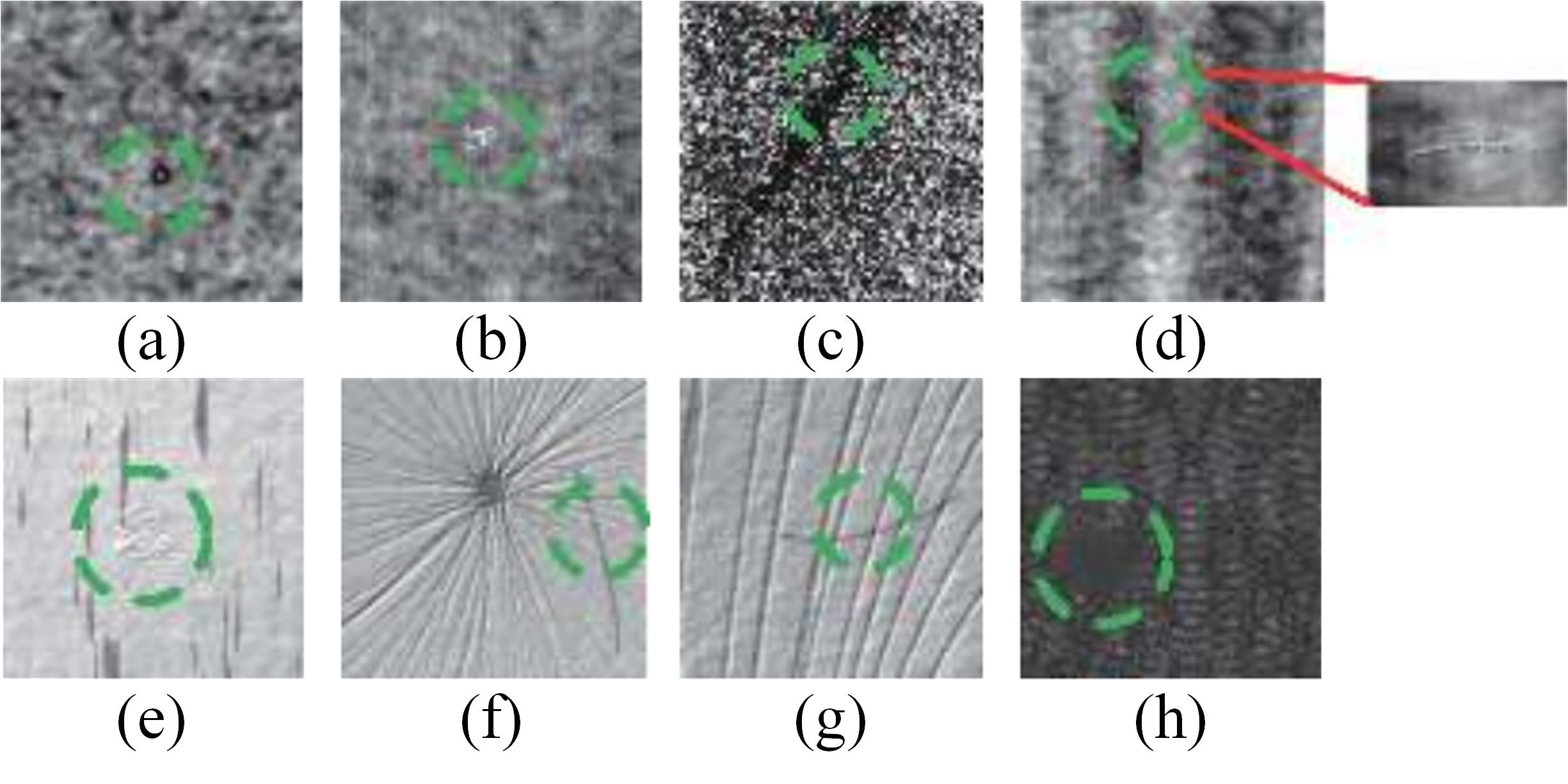}
	\caption{\centering{Various surface defects}}
	\label{fig:1}
\end{figure}

Traditional machine learning methods can effectively solve various industrial product quality detection problems, such as bearings\textsuperscript{\cite{ref1}}, mobile phone screens\textsuperscript{\cite{ref2}}, coils\textsuperscript{\cite{ref3}}, rails\textsuperscript{\cite{ref4}}, steel beams\textsuperscript{\cite{ref5}}, etc. Such methods use artificial feature extractors to adapt to a specific product image sample data set, the characteristics of the classifier and support vector machine\textsuperscript{\cite{ref6}}, and the neural network\textsuperscript{\cite{ref7}} to determine whether the product is defective. However, when the surface defects of the detected product appear, such as complex background texture (including regular and irregular), large changes in the scale of defect features, and similar defect area features and background features (as shown in Figure \ref{fig:1}), the traditional machine learning method relies on the ability of artificial features to represent product image samples. It does not adapt to such complex detection needs. Figure \ref{fig:1}(a) is a dark defect, and Figure \ref{fig:1}(b) is a light defect. Figure \ref{fig:1}(c) is a large-scale defect covering an image. Figure \ref{fig:1}(d) is a minor defect. Figure \ref{fig:1}(f)~(g) shows a defect similar to a texture. Figure \ref{fig:1}(d) shows a defect with small chromatic aberration. Figure \ref{fig:1}(h) is a fuzzy defect.

Since AlexNet\textsuperscript{\cite{ref8}} was proposed, deep learning methods based on convolutional neural networks (CNNs) have become the mainstream methods\textsuperscript{\cite{ref9-ref12}} in surface defect detection. convolutional neural networks can automatically learn image features and extract more abstract image features through the superposition of multiple convolutional layers, which has better feature representation capabilities than manually designed feature extraction algorithms. According to the results of the network output, the algorithm of defect detection by deep learning method can be divided into defect classification method, defect identification method, and defect segmentation method.

Defect classification algorithms are usually trained using classic classification network algorithms to treat the detected samples, and the learned model can classify defects and non-defect categories. Such methods usually do not involve the localization of defect areas. Wang et al.\textsuperscript{\cite{ref13}} proposed using 2 CNN networks for defect detection of 6 types of images. Xu et al.\textsuperscript{\cite{ref14}} proposed a CNN classification network that integrates visual geometry (VGG) and residual networks to detect and classify roll surface defects. Paolo et al.\textsuperscript{\cite{ref15}} and Weimer et al.\textsuperscript{\cite{ref16}} also use CNN's image feature representation capabilities to identify defects.

\section{Related Work}\label{sec2}
In order to achieve accurate localization of defect areas, some researchers have improved and applied the excellent network in vision recognition tasks to surface defect detection. Such algorithms are primarily based on regional convolutional neural networks\textsuperscript{\cite{ref17}}, single-excitation multi-box detectors\textsuperscript{\cite{ref18}}, and You only look once (YOLO)\textsuperscript{\cite{ref19}} and other networks. Chen et al.\textsuperscript{\cite{ref20}} applied deep convolutional neural networks to fastener defect detection. Cha et al.\textsuperscript{\cite{ref21}} used regional convolutional neural networks (Region-CNN, R-CNN) for structural visual inspection in construction.

In order to achieve pixel-level detection accuracy, some researchers have used segmentation networks, such as Huang et al., to convert defect detection tasks into semantic segmentation tasks, which improves the accuracy of tile surface detection. Qiu et al. use a full convolutional network (FCN) to detect defect areas.

In many industrial situations, product defect types are unpredictable and occur only during the production process, making it difficult to collect a large number of defect samples. In response to these problems, researchers began focusing on small or unsupervised samples. Such methods all rely on a certain amount of training data. Various methods improved based on autoencoders are used for surface defect detection, such as convolutional autoencoder (CAE)\textsuperscript{\cite{ref25}}, stacked noise-canceling autoencoders\textsuperscript{\cite{ref26}} based on Fisher guidelines, robust autoencoders\textsuperscript{\cite{ref27}}, sparse denoising self-coding networks\textsuperscript{\cite{ref28}} that fuse gradient difference information, etc. Xi methods, such as Yu et al.\textsuperscript{\cite{ref24}}, use the YOLOV3 network to achieve high-accuracy detection results under the training conditions of a few defective samples. Mei et al.\textsuperscript{\cite{ref29}}. proposed that a multi-scale convolutional denoising autoencoder network (MSCDAEtextsuperscript{\cite{ref29}}) reconstructs images and generates detection results using reconstruction residuals, compared with traditional unsupervised algorithms such as Phase only transform (PHOT\textsuperscript{\cite{ref30}}) and discrete cosine transformç, MSCDAE There is an excellent improvement in the model evaluation index. Yang et al\textsuperscript{\cite{ref32}}. improved the reconstruction accuracy of texture backgrounds by using feature clustering based on MSCDAE. The above-reconstructed networks all use the Mean square error (MSE) loss function with normal terms, and the data samples are mostly regular surface textures.

In addition to autoencoders, generative adversarial networks\textsuperscript{\cite{ref33}}. are also used for unsupervised defect detection. By learning a large number of normal image samples, the generator in the network can learn the data distribution of normal sample images. Zhao et al.\textsuperscript{\cite{ref34}}. combine generative adversarial networks and autoencoders, make defects on defect-free samples, and train generative adversarial networks to have the ability to recover images. He et al.\textsuperscript{\cite{ref35}}. used semi-supervision to generate anti-network and self-encoder training unlabeled steel surface defect data, extracted fine-grained features of the image, and classified them. Schlegl et al.\textsuperscript{\cite{ref36}}. propose anomaly detection to generate anomaly detection of lesion images under unsupervised conditions by generating adversarial networks. In practice, generative adversarial networks also have problems such as unstable performance and difficulty\textsuperscript{\cite{ref37}}. in training.

Considering the complexity and scarcity of defect samples in industrial application scenarios, this paper proposes a reconstituted network detection method (ReNet-D) based on a reconstructed network, which uses a small number of defect-free samples as objects for network model learning. The reconfiguration training of the sample image enables the network to have the ability to reconstruct the positive sample. When the abnormal sample is input, the trained network model can detect the abnormal area of the sample image. In this paper, the network structure, training block size, loss function coefficient, and other influencing factors of the ReNet-D method are analyzed and evaluated in detail to adapt to the detection requirements of regular textures and irregular textures time and compare experiments with other classical algorithms.

\section{Methodology}\label{sec3}

\begin{figure}[htbp]
    \centering 
    \includegraphics[width=0.9\textwidth]{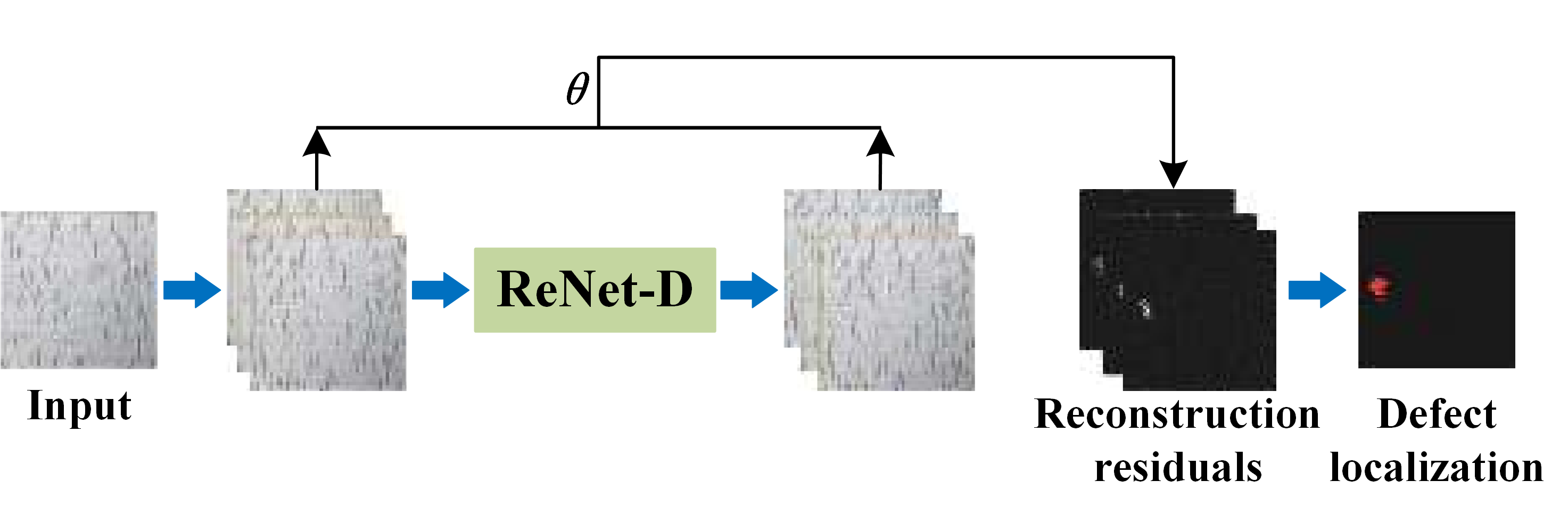}
	\caption{\centering{ReNet-D algorithm architecture}}
	\label{fig:2}
\end{figure}

In practical industrial applications, factors such as the scarcity of defective samples, significant differences in characteristics, and the accidental appearance of unknown defects make it challenging to apply supervised algorithms driven by large data samples. The unsupervised algorithm proposed in this paper solves the problem of missing insufficient sample data for model learning. The algorithm is divided into two training stages: the image reconstruction network training stage and the surface defect area detection stage. The reconstruction network is designed by the full convolutional autoencoder, and only a small number of normal samples are used for training so that the reconstruction network can generate defect-free reconstruction images. The defect detection stage takes the residuals of the reconstructed image and the image to be tested as the possible areas of defects, and the final detection results are obtained through conventional image operation. The model of the ReNet-D algorithm is shown in Figure \ref{fig:2}.

\subsection{Reconstruction Network}\label{subsec3.1}
The surface defects of industrial products have multi-scale characteristics, similar to the background texture and complex shape, which require high accuracy and operation time of the detection algorithm. Therefore, there are three requirements for refactoring the network design: 1) the network can adapt to defect areas of different scales; 2) The network needs to identify whether there are defective features in the sample area; 3) Reconstruct the network model with as few parameters as possible.

The reconstruction process is usually decomposed into encoding transforms $\varphi$ and decoding transforms $\gamma$, defined as follows:
\begin{equation}
\begin{aligned}
\varphi&: I \to F\\
\gamma&: F \to I\\
(\varphi, \gamma) &= \arg\min{\|I-\gamma(\varphi(F))\|}^2
\end{aligned}
\label{equation:1}
\end{equation}

In Eq. \eqref{equation:1}, $\mathit{I}\in\mathbf{R}^{\mathit{W}\times\mathit{H}}$ represents the spatial domain of the image sample, mapped to the hidden space by function $\varphi$, $\mathit{F}$ represents the corresponding image sample feature in the hidden space, and $\varphi$ is implemented by the coding module. $\gamma$ remaps the image sample feature $\mathit{F}$ corresponding to the hidden layer space back to the spatial domain of the original image sample, which is implemented by the decoding module.

Where $\mathit{z}=\varphi(\mathit{I})\in\mathit{F}$, the encoding and decoding process is described as:
\begin{equation}
\begin{aligned}
z&=\sigma(W\circ I + b)\\
I^{\prime}&=\sigma(W^{\prime}\circ z + b^{\prime})
\end{aligned}
\label{equation:2}
\end{equation}

In Eq. \eqref{equation:2}, $\mathit{I}^{\prime}$ represents a refactored image, $\circ$ represents convolution, $\sigma$ signifies an activation function. $\mathit{W}$ and $\mathit{W^{\prime}}$ represent the encoding convolutional kernel and the decoding convolutional kernel, respectively. $\mathit{b}$ and $\mathit{b^{\prime}}$ indicate the encoding bias and the decoding bias.

\begin{figure}[htbp]
    \centering 
    \includegraphics[width=0.9\textwidth]{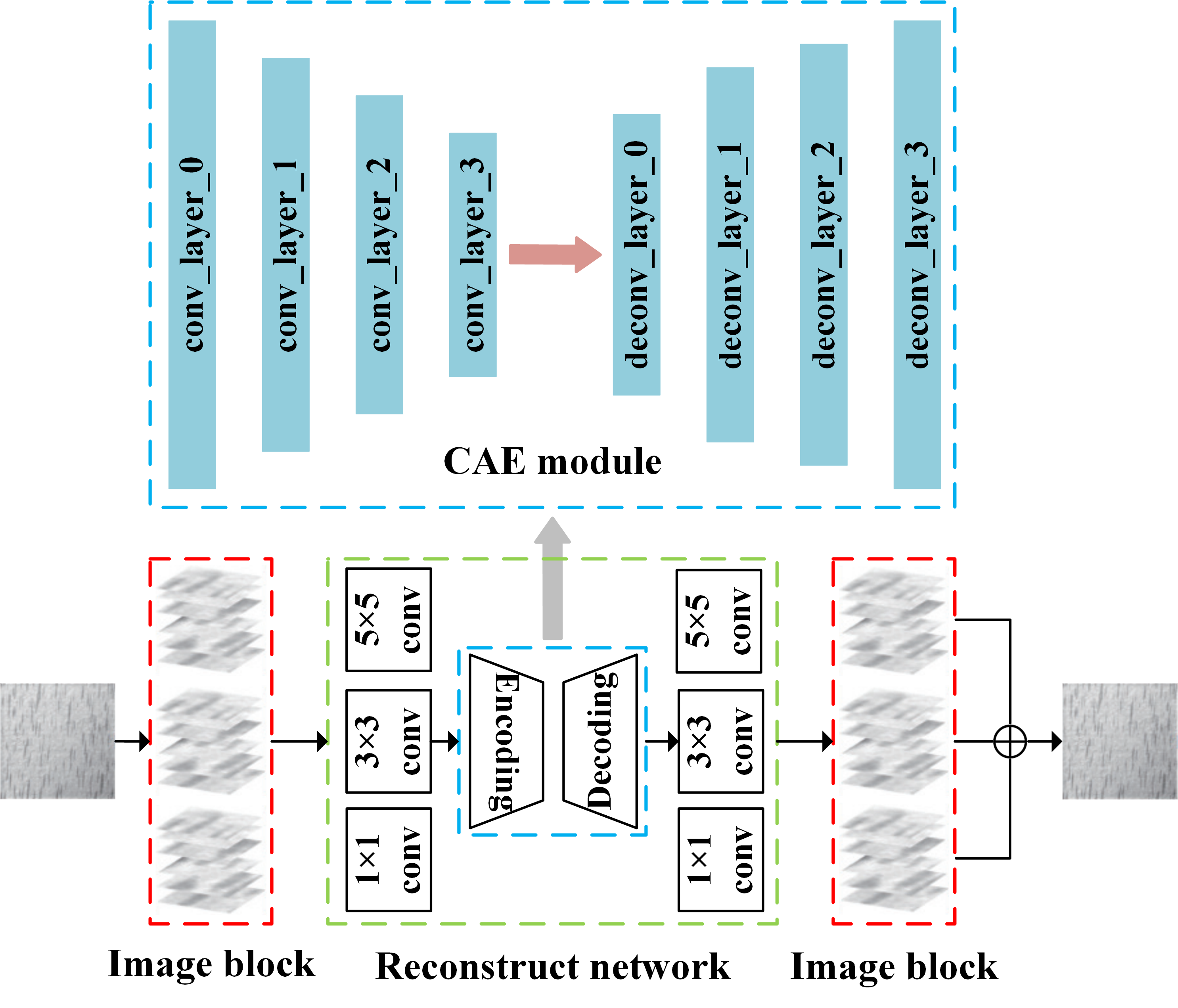}
	\caption{\centering{ReNet-D network structure}}
	\label{fig:3}
\end{figure}

The network structure of ReNet-D is shown in Figure \ref{fig:3}. In order to adapt to a larger image, the original image is divided into several image blocks, usually $16\times 16$, $32\times 32$ and $64\times 64$, as inputs to the network. ReNet-D uses the three convolutional kernels of $1\times 1$, $3\times 3$ and $5\times 5$ to obtain multi-scale features, and inputs the multi-scale features into the coding module. The results of the decoding module output are fed into three deconvolution layers of different scales to obtain the final reconstruction image, which can also obtain multi-scale features compared with the Gaussian pyramid sampling model of MSCDAE, but the computational cost is reduced. ReNet-D's CAE module consists of 4 convolutional modules and 4 deconvolution modules. Each convolutional module contains a convolutional layer, a batch normalization (BN\textsuperscript{\cite{ref38}}) layer and a nonlinear activation layer, and the first three convolutional modules also include a pooling layer that can change the image scale. The activation function uses Relu. The first 3 convolutional layers use $5\times 5$ convolutional kernels, and the last layer uses $3\times 3$ convolutional nuclei.

The mechanism of the self-encoder for defect detection is to complete the reconstruction of the defect-free background based on the high sensitivity to the defect-free environment, and to complete the imperfect reconstruction of the low sensitivity of the defective area, to realize the detection of defects. The autoencoder is a neural network with the same input and learning objectives. The depth of the network level also determines the reproduction ability of the autoencoder to the input image. If the model with a complex network structure is adopted, the ability to reproduce the sample features can be improved. Still, it also causes the reproduction ability of the defect area to be improved. ReNet-D adopts a lightweight structural design and limited capacity for model reconstruction. Still, through the design of multi-scale features and loss functions, the network can fully learn the characteristics of normal textures and obtain information on defect parts for the imperfect reconstruction of defect areas.

\subsection{Loss Function}\label{subsec3.2}
In the training phase of ReNet-D, the reconstruction error between the original and reconstructed images is used as a loss function to promote network convergence. The following analyzes and improves the existing loss function for evaluating reconstruction errors.

\textbf{Mean Squared error loss.} The textured background of most industrial products is irregular, the abnormal features are easy to integrate into the textured background, and the difference between the abnormal features and the normal texture background features is small. Define $\mathit{L}$2 as the mean absolute error loss:
\begin{equation}
L2=\|I_{src}-I_{rec}\|^2_2 + \lambda\|\omega\|_F
\label{equation:3}
\end{equation}

In the Eq. \eqref{equation:3}, $\mathit{I}_{src}$ represents the original image of the input, $\mathit{I}_{rec}$ represents the image of the model reconstruction, $\omega$ represents the set of weights in the reconstruction network, and $\lambda$ represents the penalty factor of the regularization term, $0<\lambda<1$. The reconstruction algorithm model with MSE as the loss function is suitable for image samples with regular texture backgrounds, such as textiles\textsuperscript{\cite{ref29,ref32}}.

\textbf{Average absolute error loss.} The textured background of most industrial products is irregular, the abnormal features are easy to integrate into the textured background, and the difference between the abnormal features and the normal texture background features is slight. Define $\mathit{L}$1 as the mean absolute error loss:
\begin{equation}
L1=\|I_{src}-I_{rec}\|_1 + \lambda\|\omega\|_F
\label{equation:4}
\end{equation}

Compared with $\mathit{L}$1 loss, $\mathit{L}$2 loss is more sensitive to outliers and overpasses significant loss errors, such as the MSCDAE\textsuperscript{\cite{ref29}} method, so ReNet-D introduces $\mathit{L}$1 loss to optimize network training.

\textbf{Structural loss.} When evaluating the effect of reconstructing the network model, $\mathit{L}$1 loss and $\mathit{L}$2 loss are compared on a pixel-by-pixel basis, and the regional structure of the image is not considered. For the detection of some irregular texture image samples, ReNet-D introduces the Structural similarity index (SSIM\textsuperscript{\cite{ref39,ref40}}) to construct the loss function so that the network can apply complex and changeable texture background samples to reconstruct better results. The SSIM loss function optimizes the model from three indicators(brightness, contrast, and structure)\textsuperscript{\cite{ref41}}, and the results reflect the image details better than the $\mathit{L}$1 or $\mathit{L}$2 loss functions. For image pairs ($\mathit{x}, \mathit{y}$) for model input and output, SSIM is defined as:
\begin{equation}
\begin{aligned}
SSIM(x,y)=(l(&x,y))^{\alpha}(c(x,y))^{\beta}(s(x,y))^{\gamma}\\
l(x,y)&=\frac{2u_{x}u_{y}+C_1}{u^2_{x}+u^2_y+C_1}\\
c(x,y)&=\frac{2\sigma_{xy}+C_2}{{\sigma}^2_{x}+{\sigma}^2_y+C_2}\\
s(x,y)&=\frac{\sigma_{xy}+C_3}{\sigma_{x}\sigma_{y}+C_3}
\end{aligned}
\label{equation:5}
\end{equation}

where $\alpha > 0, \beta > 0, \gamma > 0$, $\mathit{l}(\mathit{x}, \mathit{y})$ represents brightness ratio, $\mathit{c}(\mathit{x}, \mathit{y})$ represents contrast comparison, and $\mathit{s}(\mathit{x}, \mathit{y})$ represents structural comparison. $\mathit{u_x}$ and $\mathit{u_y}$ are the mean of $\mathit{x}$ and $\mathit{y}$, respectively, and $\sigma_{\mathit{x}}$ and $\sigma_{\mathit{y}}$ are the standard deviations of $\mathit{x}$ and $\mathit{y}$. $\sigma_{\mathit{xy}}$ is the covariance of $\mathit{x}$ and $\mathit{y}$. $\mathit{C}_1$, $\mathit{C}_2$,$\mathit{C}_3$ are non-zero constants, usually $\alpha = \beta = \gamma = 1$, $\mathit{C}_3 = \mathit{C}_2/2$.

The loss function of SSIM is defined as:
\begin{equation}
L_{SSIM}(x,y)=1-SSIM(x,y)
\label{equation:6}
\end{equation}

The SSIM loss function is used to evaluate the difference between the output result of the last layer and the original image in the reconstruction network, and multiple deconvolutional layer results of different scales can be extracted and the corresponding convolutional layer results can be used at the same time to use the SSIM loss function to construct multi-scale SSIM\textsuperscript{\cite{ref42}} (Multi-scale SSIM, MS\_SSIM). For M scales, the MS\_SSIM loss function is defined as:
\begin{equation}
L_{MS\_SSIM}(x,y)=1-\prod SSIM(x,y)
\label{equation:7}
\end{equation}

\textbf{Loss Function of ReNet-D}. Compared with MSE loss, the $\mathit{L}1$ loss has a weaker penalty for pixel-level error, which is suitable for irregular texture samples. At the same time, $\mathit{L_{SSIM}}$ can train the reconstruction network to pay attention to the brightness change and color deviation of the sample image to retain the high-frequency information of the image, that is, the image edge and detail. To solve the problem of defect detection of regular and random texture image samples at the same time, this paper designs a loss function combining $\mathit{L}1$ loss and $\mathit{L_{SSIM}}$ as the loss function of the ReNet-D network model, as follows:
\begin{equation}
\begin{aligned}
L_{ReNet-D}=\alpha L1 + (1-\alpha)L_{SSIM}
\end{aligned}
\label{equation:8}
\end{equation}

In the Eq. \eqref{equation:8}, $\alpha$ is the weight factor, and the value range is $(0, 1)$, which is used to balance the proportion of $\mathit{L}1$ loss and $\mathit{L_{SSIM}}$, and this paper will experimentally compare the influence of different weight factors and loss functions on the ReNet-D detection results.

\subsection{Defect Area Locate}\label{subsec3.3}
In the detection stage, the network will output an approximate defect-free image after the defect image input is trained to reconstruct the network. That is, the reconstructed network will "repair" the defective area into a normal area while maintaining the defect-free area. According to this characteristic, the pixel-level difference between the output image and the input image, after conventional image processing technology, can accurately locate the defect area, the processing process is as follows:

\begin{itemize}
\item Residual plot acquisition:
The input image (shown in Figure \ref{fig:4}(a)) and the ReNet-D reconstruction image (shown in Figure \ref{fig:4}(b)) are used to make a difference shadow to obtain the reconstruction error of the network for the defective area, and the obtained residual map is shown in Figure \ref{fig:4}(c), which contains the position information of the abnormal area. Among them, Fig. \ref{fig:4}(a) is the original image of the input model, Fig. \ref{fig:4}(b) is the ReNet-D reconstruction diagram, Fig. \ref{fig:4}(c) is the residual graph $\mathit{v}$, and ($\mathit{i}$, $\mathit{j}$) in Eq. \eqref{equation:9}. Fig. \ref{fig:4}(d) is the residual map filtering, and Fig. \ref{fig:4}(e) is the defect location.
\begin{equation}
\begin{aligned}
v(i, j)=(I_{src}(i, j)-I_{rec}(i, j))^2
\end{aligned}
\label{equation:9}
\end{equation}

\item Denoising processing:
Fig. \ref{fig:4}(c) of the residual plot shows a lot of noise, forming pseudo-defects, affecting the judgment of the real defect area, and using mean filtering for denoising to obtain Fig. \ref{fig:4}(d).
\item Thresholding segmentation and defect localization.

\begin{figure}[htbp]
	\centering
     \includegraphics[width=0.85\textwidth]{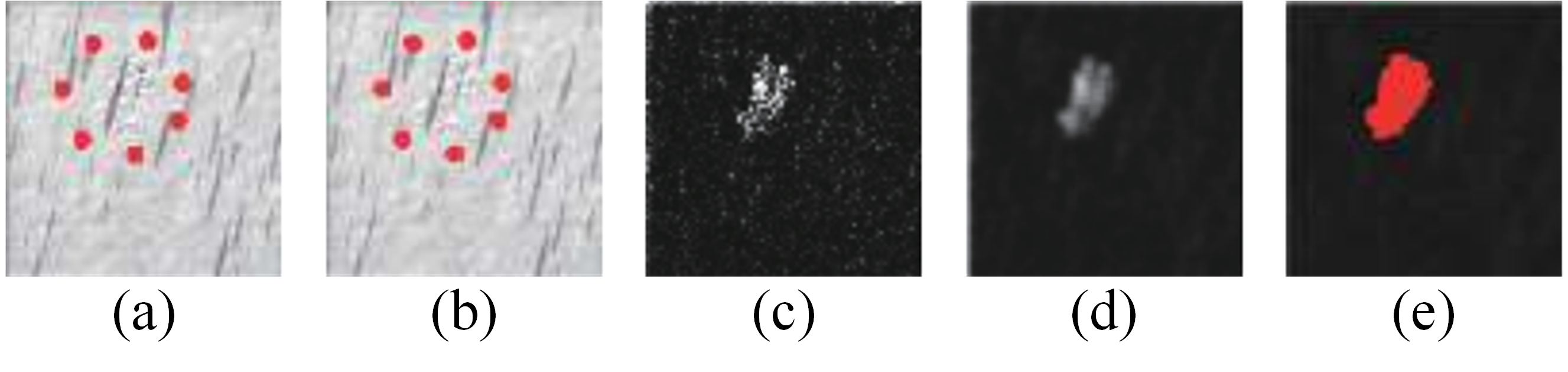}
	\caption{\centering{The residual graph processing flow location}}
	\label{fig:4}
\end{figure}

\item The final result is obtained using the adaptive threshold method Figure \ref{fig:4}(e).
\end{itemize}

\section{Experiments}\label{sec4}
In this paper, the proposed detection algorithm ReNet-D is extensively evaluated on the surface data of industrial products. Firstly, the data set used in the experiment is introduced, and then the critical indicators of model evaluation are introduced, and then the influencing factors of the detection effect of the ReNet-D algorithm include loss function, network structure, image block, and the detection effect of different types of defects of similar materials are analyzed in detail. Finally, the proposed detection algorithm is compared with other parallel unsupervised algorithms.

\subsection{Dataset}\label{subsec4.1}
\begin{figure}[htbp]
	\centering
     \includegraphics[width=0.8\textwidth]{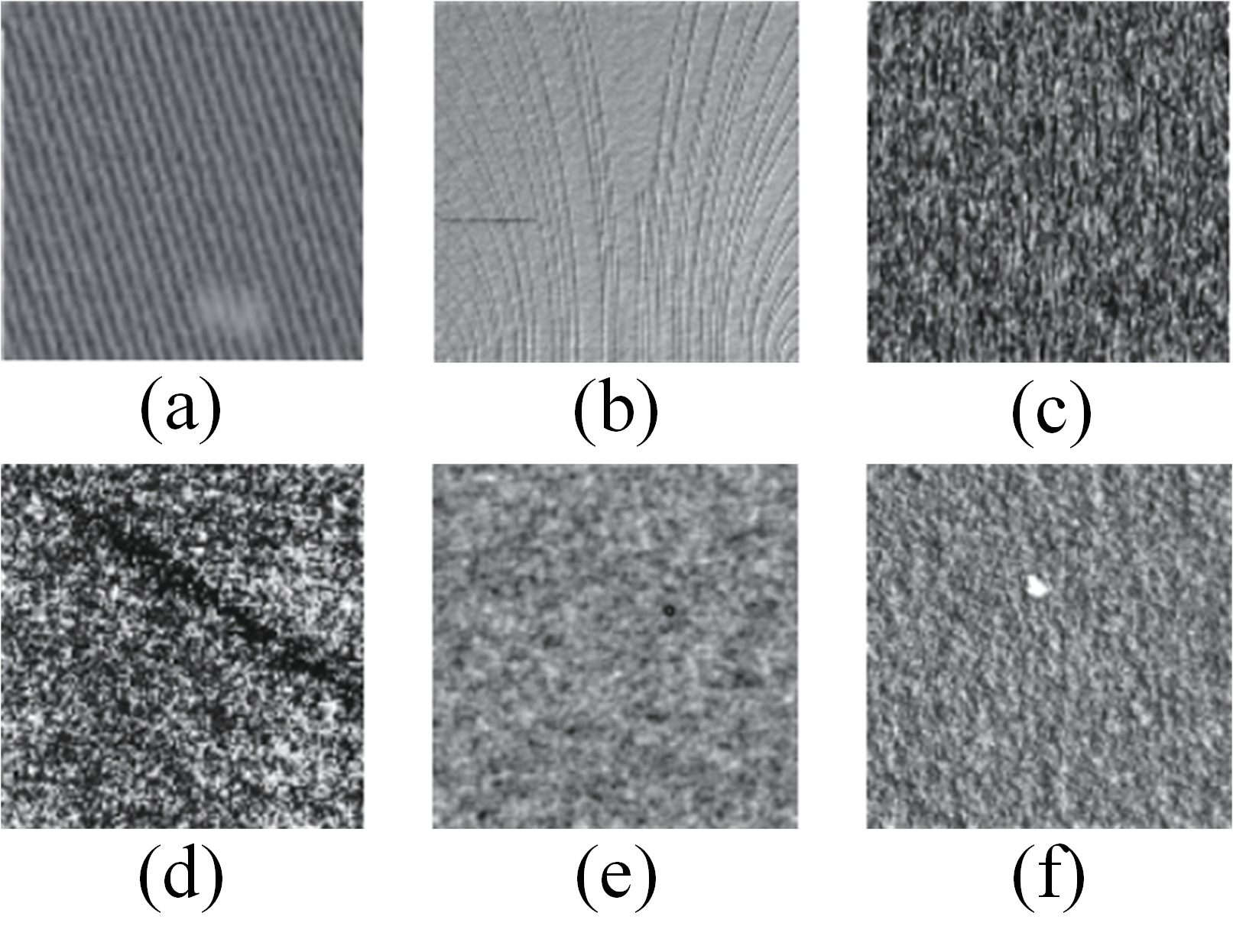}
	\caption{\centering{Surface defect data set used in the experiment}}
	\label{fig:5}
\end{figure}

To objectively evaluate the proposed detection algorithm, this experiment establishes a verification data set composed of texture samples of various materials, as shown in Figure \ref{fig:5}, where Figure \ref{fig:5}(a) is derived from the AITEX\textsuperscript{\cite{ref43}} dataset, which is derived from the textile industry, the sample is a regular texture, the number of positive and negative examples is 149/5, and Figure \ref{fig:5}(b) ~ (e) samples are derived from the DAGM2007\textsuperscript{\cite{ref44}} data set, the data set has two characteristics: texture irregularity and defect area and picture scale, defects are hidden in the texture and structure, and texture the theory is very similar, where the number of positive and negative samples in Figure \ref{fig:5}(b) is 100/29, the number of positive and negative samples in Figure \ref{fig:5}(c) is 100/6, and the number of positive and negative samples in Figure \ref{fig:5}(d) is 101/6. Figure \ref{fig:5}(f) samples from the Kylberg Sintorn dataset, plus or minus sample size 50/5. In addition to the dataset shown in Figure \ref{fig:5}, samples of the MVtech\textsuperscript{\cite{ref37}} unsupervised dataset were added for comparative experiments.

\subsection{Model Evaluation Metrics}\label{subsec4.2}
This paper evaluates the performance of the algorithm through pixel-level metrics, using three evaluation indicators: recall rate (Recall), precision rate (Precision) and the weighted harmonic average of the two ($\mathit{F}1$-Measure), which are defined as follows:
\begin{equation}
\begin{aligned}
Recall&=\frac{TP_p}{TP_p + FN_p}\times 100\%\\
Precision&=\frac{TP_p}{TP_p + FP_p}\times 100\%\\
F1\text{-}{Measure}&=\frac{2\times Precision\times Recall}{Precision + Recall}
\end{aligned}
\label{equation:10}
\end{equation}

Eq. \eqref{equation:10}, $\mathit{TP_p}$ is the proportion of correctly segmented defect areas in the foreground, $\mathit{FP_p}$ is the proportion of incorrectly segmented defect areas in the background, and $\mathit{FN_p}$ is the proportion of defect areas that are not detected in the defect area. $\mathit{F}1$-\textit{Measure} evaluates recall and precision. All tests were performed on a computer equipped with a graphics processor, as shown in Table \ref{tabular:table1}.
\begin{table}[!ht]
\caption{\label{tabular:table1}\centering{Computer system configuration}}
\centering
\setlength{\tabcolsep}{8mm}
%\resizebox{\linewidth}{!}
{\begin{tabular}{cc}
\toprule
System & Ubantu 16.04\\
\midrule
Memory   &    128 GB  \\
Gaphics Processor    &    NVIDIA GTX-1080 Ti  \\
CPU    &    Intel E5-2650 v4@2.2 GHz  \\
Deep Learning Framework    &    Pytorch, CUDA 9.0, CUDNN 5.1 \\
\botrule
\end{tabular}}
\end{table}

\subsection{Loss Function Comparison Experiment}\label{subsec4.3}
ReNet-D selects the loss functions MSE, $\mathit{L}$1, SSIM and the combination of the three for comparative experiments to evaluate the performance of the loss function proposed by Eq. \eqref{equation:8} in the defect detection task. In this experiment, the ReNet-D network parameters are set As shown in table \ref{tabular:table2}.
\begin{table}
\caption{\label{tabular:table2}\centering{Default network parameters}}
\centering
\setlength{\tabcolsep}{15mm}
%\resizebox{0.8\linewidth}{!}
{\begin{tabular}{cc}
\toprule
Block Size & 32$\times$32\\
\midrule
Number of samples   &   256  \\
Iteration steps    &    1000  \\
Loss weights $\alpha$    &    0.15 \\
\botrule
\end{tabular}}
\end{table}

\begin{figure}[htbp]
	\centering
     \includegraphics[width=0.85\textwidth]{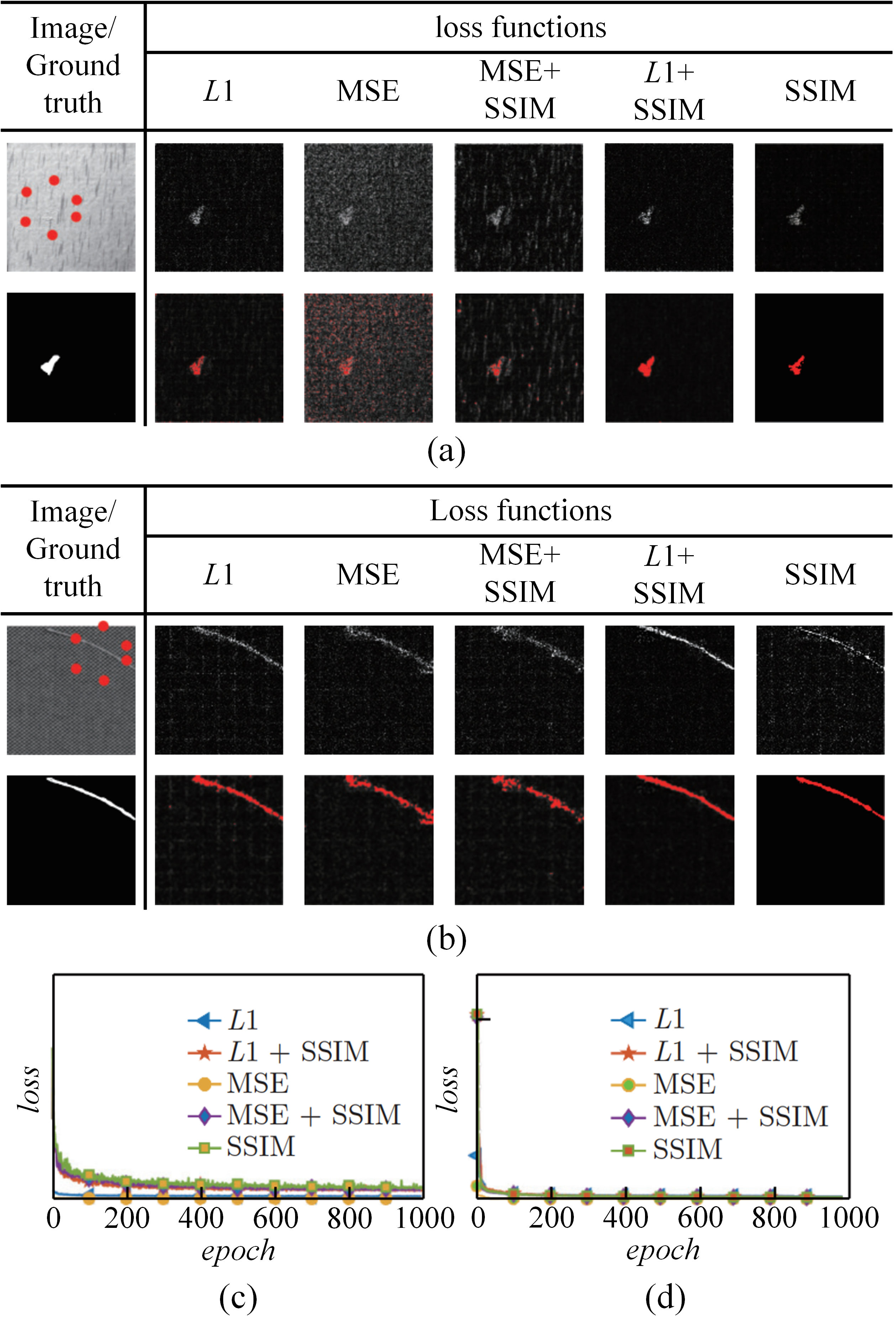}
	\caption{\centering{ReNet-D detection results under different loss functions}}
	\label{fig:6}
\end{figure}

Figure \ref{fig:6}(a) and Figure \ref{fig:6}(b) are the experimental results of two different product surface defect samples under various loss functions, of which Figure \ref{fig:6}(a) is an irregular texture sample and Figure 6(b) is a regular texture sample. Figure \ref{fig:6}(c) is the convergence test of sample (a) under different loss functions, Figure \ref{fig:6}(d) is the convergence test of sample (b) under different loss functions, and Figure \ref{fig:6}(a) uses an irregular surface texture Defect image sample. From the comparison of residual results, it can be seen that the residual results obtained by MSE as the loss function of the algorithm have more noise points in other areas except the actual defect area, forming false defects; while using SSIM alone as the loss function, the detection The defect area is slightly smaller than the actual defect area; compared with other loss functions, the combination of the structural loss function SSIM and the $\mathit{L}1$ loss function achieves better results.

Figure \ref{fig:6}(b) Defect image sample with regular surface texture. From the comparison of residual results, it can be seen that the integrity of the defect area obtained by using the MSE loss function is poor, similar to the detection result of the MSE + SSIM. Structural loss function The combination of SSIM and $\mathit{L}1$ loss function achieves better results, and the detection results are similar to using only the $\mathit{L}1$ loss function. Figure \ref{fig:6}(c) and Figure \ref{fig:6}(d) compare the convergence trends of ReNet-D under different loss functions.

\begin{table*}[!ht]
\caption{\label{tabular:table3}\centering{Comparison of test results under different loss functions}}
\centering
\resizebox{\linewidth}{!}
{\begin{tabular}{ccccccc}
\toprule
Metrics & Samples & $\mathit{L}1$ & MSE & MSE + SSIM & $\mathit{L}1$ + SSIM & SSIM \\
\midrule
\multirow{2}{*}{Recall} & irregular samples & 0.51 & 0.38 & 0.5 & \textbf{0.75} & 0.59 \\
& regular samples & \textbf{0.76} & 0.70 & 0.67 & 0.71 & 0.59 \\
\midrule
\multirow{2}{*}{Precision} & irregular samples & \textbf{0.93} & 0.35 & 0.52 & 0.89 & \textbf{0.93} \\
& regular samples & 0.84 & 0.65 & 0.70 & 0.87 & \textbf{0.96} \\
\midrule
\multirow{2}{*}{Weights} & irregular samples & 0.66 & 0.36 & 0.51 & \textbf{0.82 } & 0.72 \\
& regular samples & \textbf{0.80} & 0.67 & 0.69 & 0.78 & 0.73 \\
\botrule
\end{tabular}}
\end{table*}

The comparison results in Table \ref{tabular:table3} show that for the defect samples with irregular surface texture in Figure \ref{fig:6}(a), the combination of the structural loss function SSIM and the $\mathit{L}1$ loss function achieves better recall and weighted harmonic mean results. It is slightly inferior to the loss function SSIM in precision. In the defect samples with regular surface texture shown in Fig. \ref{fig:6}(b), only the use of the $\mathit{L}1$ loss function achieves the highest recall rate, followed by the combination of the SSIM and $\mathit{L}1$ loss functions. The loss function SSIM achieves the highest accuracy, followed by the variety of SSIM and $\mathit{L}1$ loss functions. For weighted harmonic averaging, the $\mathit{L}1$ loss function performs the best.

Further, by comparing the detection effects of the sample library used in this study, the loss function has the following rules:
\begin{itemize}
\item[1)]  For regular surface texture samples, the above four loss functions can be used to detect defects, among which the results of using only the MSE and MSE + SSIM loss functions are relatively poor, and the results of the other two loss functions are slightly different.
\item[2)] For irregular surface texture samples, the detection results obtained using the combined loss function of $\mathit{L}1$ + SSIM are better.
\end{itemize}

\begin{figure}[htbp]
	\centering
     \includegraphics[width=0.85\textwidth]{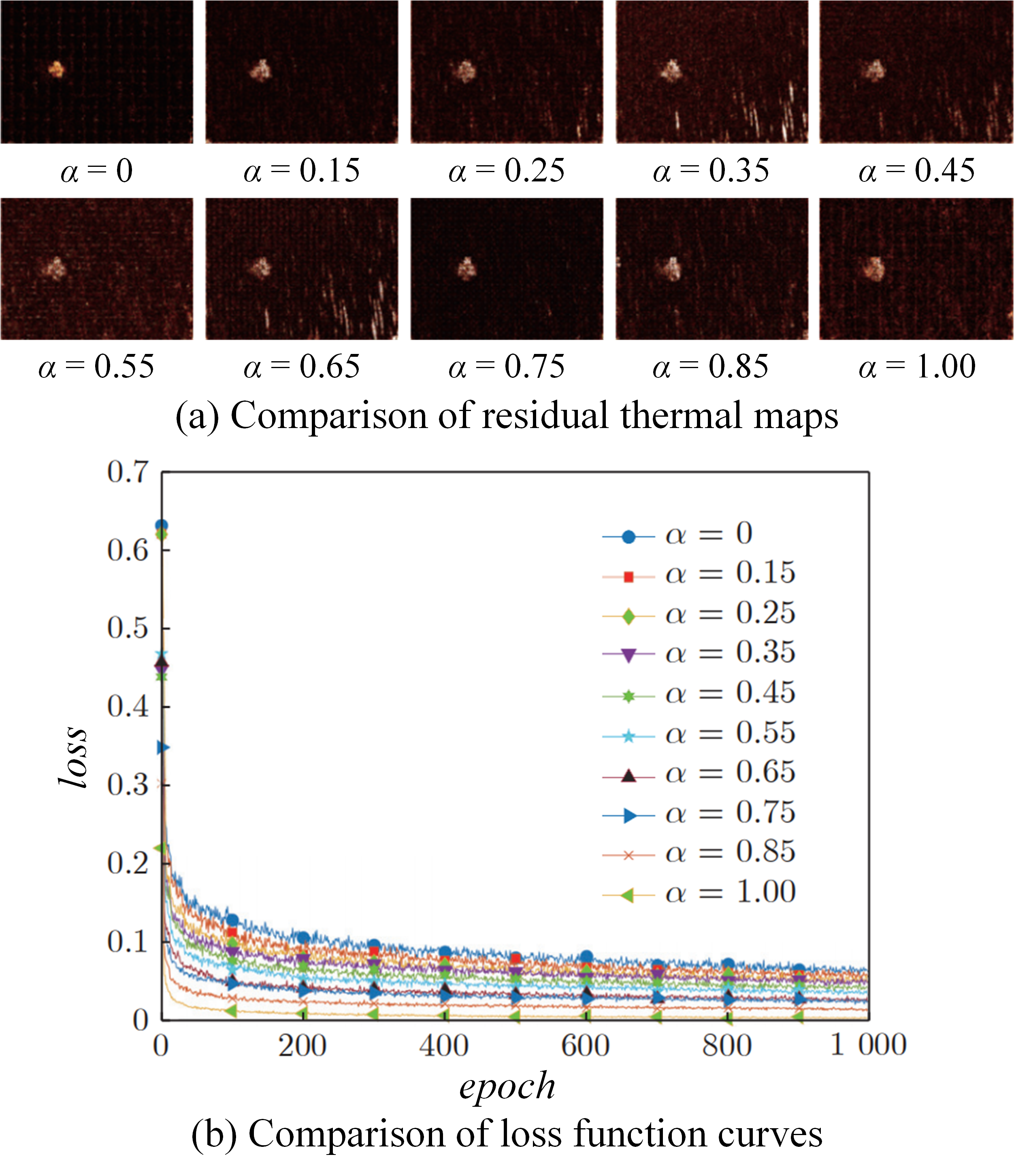}
	\caption{\centering{Comparison of ReNet-D performances under different weight coefficients}}
	\label{fig:7}
\end{figure}

\begin{table*}[!ht]
\caption{\label{tabular:table4}\centering{Comparison of test results under different loss functions}}
\centering
\resizebox{\linewidth}{!}
{\begin{tabular}{ccccccccccc}
\toprule
Weights & 0 & 0.15 & 0.25 & 0.35 & 0.45 & 0.55 & 0.65 & 0.75 & 0.85 & 1 \\
\midrule
Recall & 0.72 & \textbf{0.79} & 0.62 & 0.73 & 0.65 & 0.67 & 0.52 & 0.55 & 0.72 & 0.45 \\
\midrule
Precision & 0.71 & 0.69 & 0.58 & 0.28 & 0.46 & 0.53 & 0.23 & \textbf{0.89} & 0.54 & 0.62 \\
\midrule
Weights & 0.71 & \textbf{0.73} & 0.60 & 0.41 & 0.54 & 0.60 & 0.32 & 0.68 & 0.62 & 0.52 \\
\botrule
\end{tabular}}
\end{table*}

\subsection{Ablation Experiments}\label{subsec4.4}
\subsubsection{Different weight coefficients}\label{subsubsec4.4.1}
The above experiments show that the ReNet-D model uses the combined loss function of $\mathit{L}1$ + SSIM, which can be applied to defect detection of both regular and irregular surface textures. For normal texture samples, only using the $\mathit{L}1$ loss, that is, the weight coefficient $\alpha = 1$, can obtain a relatively For irregular texture samples, different weight coefficients will produce different detection effects. This experiment uses the irregular texture samples in Figure \ref{fig:6}(a). The weight coefficient $\alpha$ ranges from 0 to 1, and the step size is set to 0.1 and is used to adjust the proportion of SSIM loss and $\mathit{L}1$ loss. The comparison experiment is shown in Figure \ref{fig:7}, in which Figure \ref{fig:7}(a) is the residual heat map comparison. Figure \ref{fig:7}(b) is the training loss curve comparison.

According to Eq. \eqref{equation:8}, when $\alpha$ increases, the influence of structural loss SSIM gradually decreases. It can be seen from Figure \ref{fig:7} and Table \ref{tabular:table4} that the residual map changes significantly. When $\alpha = 0.15$, the defect detection effect is better, the signal-to-noise ratio is the lowest, and the recall rate and weighted harmonic average are the best. Through experiments with multiple samples, the practical advice given in this paper is for regular texture samples to set the weight $\alpha$ = 1, which only uses the $\mathit{L}1$ loss as the training model. The loss function of irregular texture samples, set $\alpha = 0.15$, so that the weight of the structural loss effect is too large to obtain the best results.

\subsubsection{Different kinds of defects}\label{subsubsec4.4.2}
\begin{figure}[htbp]
	\centering
     \includegraphics[width=0.9\textwidth]{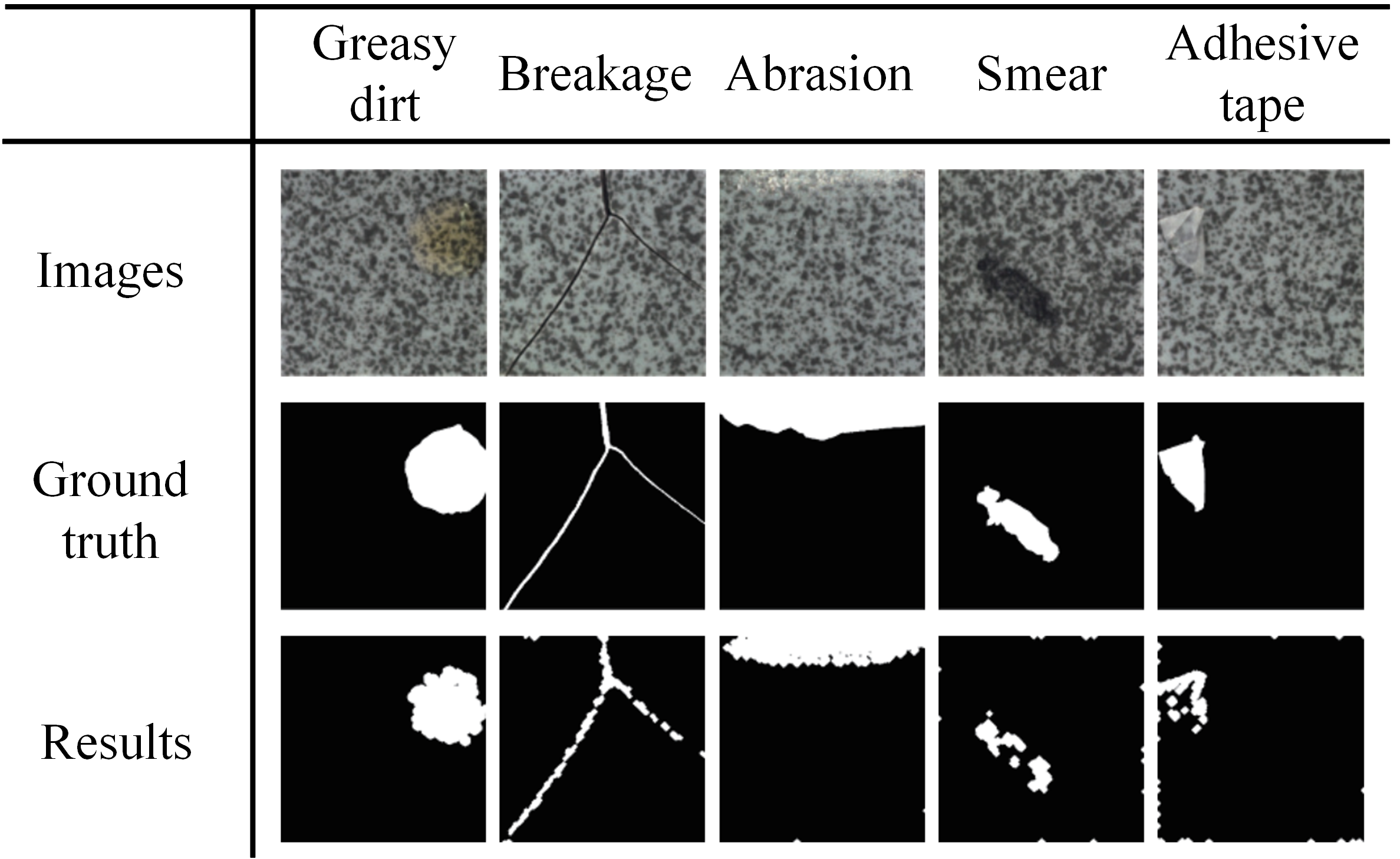}
	\caption{\centering{Test results of unsupervised samples}}
	\label{fig:8}
\end{figure}

The surface texture defect classes in the unsupervised dataset include leather, wood, carpet, grid and tiles, among which the tile class has the most messy textures. The training set includes 230 non-defective irregular texture normal type images; the test set includes 5 There are 17 damage defects, 18 tape defects, 16 gray smear defects, 18 oil stain defects, and 15 wear scar defects. In this experiment, the default network parameters are used to perform the Tile data set. After training, the detection results of 5 types of defects are obtained as shown in Figure \ref{fig:8}.

As shown in Table \ref{tabular:table5}, the ReNet-D method can adapt to the detection of different types of defects in the Tile data set, among which oil stains, damage and wear scars have better performance, and its detection precision, recall and weighted harmonic average are better , but in the detection of smear and tape defects, the defect area with different color and texture from the background can be detected, but the overall shape of the generated defect is different from the ideal detection result, resulting in a low pixel-level evaluation index.
\begin{table}[!ht]
\caption{\label{tabular:table5}\centering{Test results of unsupervised samples}}
\centering
\setlength{\tabcolsep}{8mm}
%\resizebox{\linewidth}{!}
{\begin{tabular}{cccc}
\toprule
Category & Recall & Precision & weights\\
\midrule
dirt   &    0.71 & 0.94  & 0.80  \\
damaged    &    0.66 & 0.48 & 0.55 \\
crack    &    0.63  & 0.89 & 0.70  \\
smear    &    0.27 & 0.47 & 0.32 \\
tape    &   0.16 & 0.35 & 0.20 \\
\botrule
\end{tabular}}
\end{table}

The characteristic of ReNet-D is that the network has a better reconstruction effect for the components similar to the background, and the reconstruction effect is poor for the parts that are not similar to the environment, so this feature can be used to detect defects that are different from the background. The analysis of the tape defect characteristics found that the local area's color is very close to the texture and environment or even coincides with it. The detection effect of the region is not affected. Although it is not ideal regarding recall rate and other indicators, in industrial detection, local detection can be regarded as defect detection.

This paper compares the efficiency of the algorithms. The experiment uses $1024\times1024$ pixel sample images. Under the same computing performance, the processing time of the four methods is compared, as shown in Table \ref{tabular:table6}. After the ReNet-D algorithm model is trained The size is less than 1 MB bytes, and the average detection time is 2.82 ms, which can meet the requirements of industrial real-time detection. Other methods are time-consuming, which limits their practical application.
\begin{table}[!ht]
\caption{\label{tabular:table6}\centering{Comparison of processing time (ms)}}
\centering
\setlength{\tabcolsep}{6mm}
%\resizebox{\linewidth}{!}
{\begin{tabular}{ccccc}
\toprule
Methods & PHOT & LCA & MSCDAE & ReNet-D\\
\midrule
Time   &  450 &  430 & 9746.59  & 2.82  \\
\botrule
\end{tabular}}
\end{table}

\section{Conclusions}\label{sec5}
This paper proposes a novel ReNet-D for the visual detection of surface defects using a reconstruction network. This method uses a fully convolutional autoencoder with a lightweight structure to design a reconstruction network. It can solve the problem of difficulty in obtaining defect samples in the industrial environment; in the detection stage, the trained model is used to reconstruct the input defect samples, and conventional image processing algorithms can accurately detect the defect area. In the training phase, only defect-free samples are used for training. This paper discusses the unsupervised influence of factors such as network structure and loss function in the algorithm on the surface defect detection task, and a combined loss function combining $\mathit{L}1$ loss and structural loss is proposed for surface defect detection to adapt to the detection problem of regular texture and irregular texture samples at the same time. This paper The proposed ReNet-D method is compared with other unsupervised algorithms on multi-class sample data. The results show that the detection algorithm proposed in this paper has achieved good results and is suitable for transplanting to industrial detection environments. Due to the lightness of the quantitative network characteristics, ReNet-D has better reconstruction performance for some defects that are similar to the background texture and close to the color, resulting in inconspicuous shadow results. The improvement makes the defect contrast more noticeable and achieves a better detection effect.

\end{document}